\documentclass[9pt,conference]{IEEEtran}
\IEEEoverridecommandlockouts
\usepackage{cite}
\usepackage{amsmath,amssymb,amsfonts}
\usepackage{amsthm}
\usepackage{algorithmic}
\usepackage{graphicx}
\usepackage{textcomp}
\usepackage[dvipsnames]{xcolor}
\usepackage{subcaption}
\usepackage{caption}
\usepackage{comment}
\usepackage{url}
\usepackage{enumitem}
\def\BibTeX{{\rm B\kern-.05em{\sc i\kern-.025em b}\kern-.08em
    T\kern-.1667em\lower.7ex\hbox{E}\kern-.125emX}}
\begin{document}

\def\x{{\mathbf x}}
\def\L{{\cal L}}
\newcommand{\ie}{i.e.,}

% Title.
% ------
\title{Higher-Order Topological Directionality and \\Directed Simplicial Neural Networks
\thanks{\hspace{-.4cm} Part of this work was supported by the TU Delft AI Labs program, the NWO OTP GraSPA proposal \#19497, and the NWO VENI proposal 222.032. M. Lecha is with the Istituto Italiano di Tecnologia, Genoa, Italy. Email: manuel.lecha@iit.it. A. Cavallo and E. Isufi are with the Delft University of Technology, Delft, Netherlands. Emails: \{a.cavallo,e.isufi-1\}@tudelft.nl F. Dominici and C. Battiloro are with Harvard University, Cambridge, MA, USA. Emails: \{fdominic,cbattiloro\}@hsph.harvard.edu
}
}
%
% Single address.
% ---------------

\author{
\IEEEauthorblockN{
Manuel Lecha,
Andrea Cavallo,
Francesca Dominici,
Elvin Isufi,
Claudio Battiloro
} %
}

% \author{
% \IEEEauthorblockN{
% Manuel Lecha\IEEEauthorrefmark{1},
% Andrea Cavallo\IEEEauthorrefmark{2},
% Francesca Dominici\IEEEauthorrefmark{3},
% Elvin Isufi\IEEEauthorrefmark{2},
% Claudio Battiloro\IEEEauthorrefmark{3}
% } %
% \IEEEauthorblockA{
% \IEEEauthorrefmark{1}Istituto Italiano di Tecnologia, Genoa, Italy } %
% \IEEEauthorblockA{
% \IEEEauthorrefmark{2}Delft University of Technology, Delft, Netherlands } %
% \IEEEauthorblockA{
% \IEEEauthorrefmark{3}Harvard University, Cambridge, MA, USA } %
% }

\newcommand{\claudio}[1]{{\color{orange}#1}}
\newcommand{\manu}[1]{{\color{black}#1}}
\newcommand{\maybe}[1]{{\color{gray}#1}}
\newcommand{\error}[1]{{\color{purple}#1}}
\newcommand{\andrea}[1]{{\color{orange}#1}}
\newtheorem{lemma}{Lemma}
\newtheorem{theorem}{Theorem}
\newtheorem{remark}{Remark}
\newenvironment{customproof}[1][Proof]{\noindent\textbf{#1.} }{\ \rule{0.5em}{0.5em}}
\renewcommand{\qedsymbol}{$\blacksquare$}

\maketitle

\begin{abstract}
Topological Deep Learning (TDL)  has emerged as a paradigm to process and learn from signals defined on higher-order combinatorial topological spaces, such as simplicial or cell complexes. Although many complex systems have an asymmetric relational structure, most TDL models forcibly symmetrize these relationships. In this paper, we first introduce a novel notion of higher-order directionality and we then design Directed Simplicial Neural Networks (Dir-SNNs) based on it. Dir-SNNs are message-passing networks operating on directed simplicial complexes able to leverage directed and possibly asymmetric interactions among the simplices. To our knowledge, this is the first TDL model using a notion of higher-order directionality. We theoretically and empirically prove that Dir-SNNs are more expressive than their directed graph counterpart in distinguishing \manu{non-isomorphic} directed graphs. Experiments on a synthetic source localization task demonstrate that Dir-SNNs outperform undirected SNNs when the underlying complex is directed, and perform comparably when the underlying complex is undirected.
\end{abstract}
\vspace{-.2cm}
\section{Introduction}
\label{sec:intro}
\vspace{-.1cm}
A strong inductive bias for deep learning models is processing signals while respecting the relational structure of their underlying space. Topological Deep Learning (TDL) is an emerging paradigm to process and learn from signals defined on combinatorial topological spaces (CTS) like simplicial and cell complexes \cite{hajij2023topological, papamarkou2024position}. Unlike traditional graphs and Graph Neural Networks (GNNs), which can capture only pairwise relationships, i.e., two nodes connected by an edge, combinatorial topological spaces and Topological Neural Networks (TNNs) can capture higher-order interactions \cite{barbarossa2020topological}. Such higher-order interactions are essential in many interconnected systems, including biological networks, where multi-way links exist among genes, proteins, or metabolites \cite{lambiotte2019networks}. 
In this work, we are interested in Simplicial Complexes, powerful CTS allowing for more sophisticated adjacency schemes among \textit{simplices} (nodes or groups of nodes closed under inclusion) than graphs, thus leading to a richer topological characterization and inductive bias. 

\textbf{Related Works.} \textit{Topological Neural Networks} have been shown to be expressive \cite{bodnar2021weisfeiler} (using the WL criterion \cite{xu2018powerful}), able to handle long-range interactions \cite{giusti2023cell}, and effective in heterophilic settings \cite{sheaf2022, battiloro2023tangent}. Simplicial convolutional neural networks have been proposed in \cite{ebli2020simplicial,yang2023convolutional}.  Message-passing simplicial networks have been introduced in \cite{bodnar2021weisfeiler} along with a simplicial generalization of the WL test. Recurrent message-passing simplicial networks were explored in \cite{roddenberry2019hodgenet}. $E(n)$ equivariant message-passing simplicial networks have been introduced in \cite{eijkelboom2023mathrmen}. Message-passing free simplicial neural networks have been proposed in \cite{gurugubelli2023simpleyetsimplicial,maggs2024simplicial}. Finally, simplicial attention networks have been proposed in \cite{battiloro2023generalized, giusti2022simplicial, anonymous2022SAT, lee2022sgat}.

 \textbf{Current Gaps.} Accounting for edge directionality \manu{\textit{has transformed}} our understanding of networks modeled as graphs and led to better learning performance \cite{Rossi23, zhang2021magnet, thost2021directedDAG, tong2020directedGCN}. We anticipate that the emerging research on directionality in higher-order networks modeled as CTS \cite{Riihimaki24qconnect, gong2024higher, fuchsgruber2024edge} will have a similar transformative impact. However, extending the TDL machinery to account for higher-order directionality remains unexplored and challenging. The main reason for this is that the very concept of higher-order directionality is not yet well defined in the Topological Deep Learning literature. 

\textbf{Contribution.} Here, we fill this gap with a three-fold contribution.
\begin{enumerate}[leftmargin=19pt]
    \item[\textit{\textbf{C1.}}] We \textbf{motivate and introduce} \textbf{a novel notion of higher-order topological directionality}, i.e., novel directed adjacencies among the simplices of a \textit{directed simplicial complex}, hinging on the theory of directed simplicial paths \cite{Riihimaki24qconnect}. Directed simplicial paths generalize directed paths among nodes to directed paths among higher-order simplices. 

     \item[\textit{\textbf{C2.}}] We introduce the first directed Topological Neural Networks: \textbf{Directed Simplicial Neural Networks} (Dir-SNNs), message-passing networks operating on directed simplicial complexes and leveraging the above directed adjacencies.

      \item[\textit{\textbf{C3.}}] We theoretically and empirically prove that directed simplicial neural networks \textbf{can distinguish \manu{non-isomorphic} directed graphs} better than directed graph neural networks.
\end{enumerate}

 We numerically illustrate the potential of Dir-SNNs on a source localization task at the edge level, confirming that both directionality and the directed topological inductive bias play a role when compared with undirected SNNs and directed GNNs, respectively. 
\vspace{-.2cm}
\section{Background}
\vspace{-.1cm}
\label{sec:background}

In this section, we define our domain of interest, \textit{directed simplicial complexes}, our main tool to define higher-order directionality, \textit{face maps}, and the signals we are interested into, \textit{topological signals}. 

\textbf{Directed Simplicial Complexes.} An (abstract) undirected simplicial complex is a pair $\widetilde{\mathcal{K}} = (V, \Sigma)$, where $V$ is a finite set of vertices, and $\Sigma$ is a collection of \manu{non-empty finite subsets of vertices such that (1) for every $v \in V$, $v\in \Sigma$ and (2) for every element $\sigma \in \Sigma$, if $\tau \subseteq \sigma$ and $\tau$ is non-empty, then $\tau \in \Sigma$ (inclusivity property) \cite{barbarossa2020topological}. An element $\sigma$ of $\Sigma$ is called a \textit{simplex} of $\widetilde{\mathcal{K}}$. Semi-simplicial sets (also known as $\Delta$-sets) generalize simplicial complexes by multiple distinct simplices to share the same set of vertices \cite{eilenberg50}. This generalization is crucial for modeling directed structures, such as directed graphs (digraphs), where directed edges like $(1,0)$ and $(0,1)$ must be treated as distinct simplices to capture asymmetry and directionality. A directed simplicial complex $\mathcal{K} = (V, \Sigma)$ is a $\Delta$-set in which $\Sigma$ is a collection of non-empty ordered tuples of vertices called \textit{directed simplices}. Again, (1) and (2) hold.} The dimension $\dim(\sigma)$ of a directed simplex $\sigma = (v_0, \dots, v_k)$ is given by its arity minus one, i.e., $\dim(\sigma)=k$ when $|\sigma|=k+1$. If $\dim(\sigma)=k$, $\sigma$ is called a $k$-simplex. The dimension  $\dim(\mathcal{K})$ of a directed simplicial complex $\mathcal{K}$ is the maximal dimension of a directed simplex in $\mathcal{K}$. The $k$-skeleton $\mathcal{K}_k$ of $\mathcal{K}$ is the collection of directed simplices of dimension up to $k$. A directed graph is an example of a directed simplicial complex of dimension one, in which nodes and directed edges are directed 0- and 1-simplices, respectively. A directed simplicial complex of dimension two also comprises directed triangles. 

\begin{figure}[t]
    \centering
    \includegraphics[width=0.85\columnwidth]{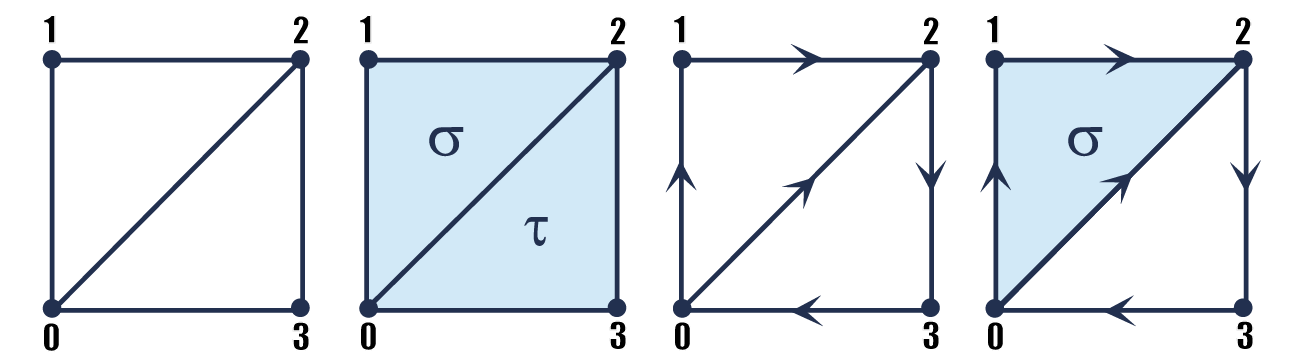} % Adjust the width as needed
    \caption{From left to right: an undirected graph, its corresponding two-dimensional flag complex (nodes, edges, and triangles), a directed graph, and its corresponding directed flag complex (nodes, directed edges, and directed triangles). In the undirected flag complex, the triangles are $\sigma = \{0,1,2\}$ and $\tau = \{0,2,3\}$. In the directed flag complex, only the triangle $\sigma = (0,1,2)$ forms a directed triangle, as $(0,2,3)$ lacks the required edge $(0,3)$.}
    \label{fig:dfgs}
\end{figure}

\textbf{Directed Flag Complexes.} It is often useful to address (directed) graph-based problems by enriching the graph with higher-order relations, thereby mapping it to a higher-order \manu{(directed)} simplicial complex while preserving its structure -- nodes (0-simplices) and edges (1-simplices) remain those of the underlying graph. When such a structure-preserving transformation maps \manu{indistinguishable} graphs to \manu{indistinguishable} simplicial complexes (formally, it preserves isomorphisms) and is injective, meaning it maps \manu{distinguishable} graphs to \manu{distinguishable} simplicial complexes, it is called a \textit{graph lifting} \cite{bodnarcwnet, bodnar2021weisfeiler}. A prominent example is the flag complex lifting, which maps an undirected graph $\mathcal{G}$ to a flag complex $\widetilde{\mathcal{K}}_{\mathcal{G}}$ \cite{chong21flag&homology}. A flag complex is a simplicial complex where the $k$-simplices correspond to the $(k+1)$-cliques in $\mathcal{G}$ -- subsets of $k+1$ vertices where each pair of distinct vertices is connected by an edge. Directed flag complex liftings extend this transformation to accommodate digraphs. In this case, a directed flag complex $\mathcal{K}_{\mathcal{G}}$ is a directed simplicial complex where the $k$-simplices are ordered $(k+1)$-cliques in $\mathcal{G}$. In this context, an ordered $k$-clique is a totally ordered tuple $(v_1, \dots, v_k)$ such that $(v_i, v_j)$ is a directed edge for $i < j$. A simple example of a graph, its flag complex, a directed graph, and its directed flag complex is presented in Fig. \ref{fig:dfgs}.

\textbf{Face Maps.} In a directed simplicial complex $\mathcal{K}$, if $\tau \subseteq \sigma$ and $\sigma \in \mathcal{K}$, then $\tau$ is said to be a face of $\sigma$. Specifically, if $\dim(\tau) = \dim(\sigma) - 1$, $\tau$ is called a facet of $\sigma$. A directed $k$-simplex has $k+1$ facets. Face maps are a formal tool used to identify the faces of a simplex by systematically removing one of its vertices. Let $\mathcal{K} = (V,\Sigma)$ be a directed simplicial complex of dimension $K$.  \manu{For $k \geq 1$}, the face map $d_i$ maps a $k$-simplex $\sigma$ to the $(k-1)$-simplex $\tau_i$ obtained by omitting the $i$-th vertex from $\sigma$, thus it is defined as:
%\begin{equation}\label{eq:face_maps} d_i(\sigma) = d_i(v_0, v_1, \dots, v_k) = (v_0, \dots, \hat{v}_i, \dots, v_k) = \tau_i,\end{equation}
\begin{equation}
\label{eq:face_map}
d_i(\sigma) = \tau_i =
\begin{cases}
(v_0, \ldots, \hat{v}_i, \ldots, v_k) & \text{if } i < k, \\
(v_0, \ldots, v_{k-1}, \hat{v}_k) & \text{if } i \geq k.
\end{cases},
\end{equation}
where $\hat{v}_i$ means that the vertex in the $i$-th position has been removed. The resulting $(k-1)$-simplex $\tau_i$ preserves the original order of vertices, excluding the omitted vertex. For instance, applying the face maps to a directed triangle returns its directed edges.  Face maps satisfy the simplicial identity $d_i \circ d_j = d_{j-1} \circ d_i$, where $\circ$ is the composition operator, for $i < j$, ensuring that the order in which vertices are removed does not affect the resulting face. Face maps are an essential tool for understanding and structuring the relationships between directed simplices within a complex. 

\textbf{Topological Signals.} Given a directed simplicial complex $\mathcal{K} = (V, \Sigma)$, a topological signal over $\mathcal{K}$  is defined as a mapping $x: \Sigma \rightarrow \mathbb{R}$ from the set of simplices $\Sigma$ to real numbers. Therefore, the \textit{feature vectors} $\mathbf{x}_\sigma \in \mathbb{R}^F$  and $\mathbf{x}_\tau \in \mathbb{R}^F$ of simplices $\sigma$ and $\tau$ are a collection of $F$ topological signals, i.e., 
\begin{equation}\label{eq:signals}
    \mathbf{x}_\sigma = [x_1(\sigma),\dots, x_F(\sigma)]^\top \textrm{ and }\mathbf{x}_\tau = [x_1(\tau),\dots,x_F(\tau)]^\top.
\end{equation}
For example, in a directed simplicial complex of dimension two, there are $F$ signals (features) associated with nodes, edges, and triangles.
\vspace{-.2cm}
\section{Directed Simplicial Neural Networks}
\label{sec:spnns}
\vspace{-.1cm}

In this section, we introduce Directed Simplicial Neural Networks (Dir-SNNs), message-passing networks operating on directed simplicial complexes. To do so, we first formally motivate and then introduce a consistent notion of higher-order topological directionality.

\begin{figure}[t]
    \centering
    \includegraphics[width=0.75\linewidth]{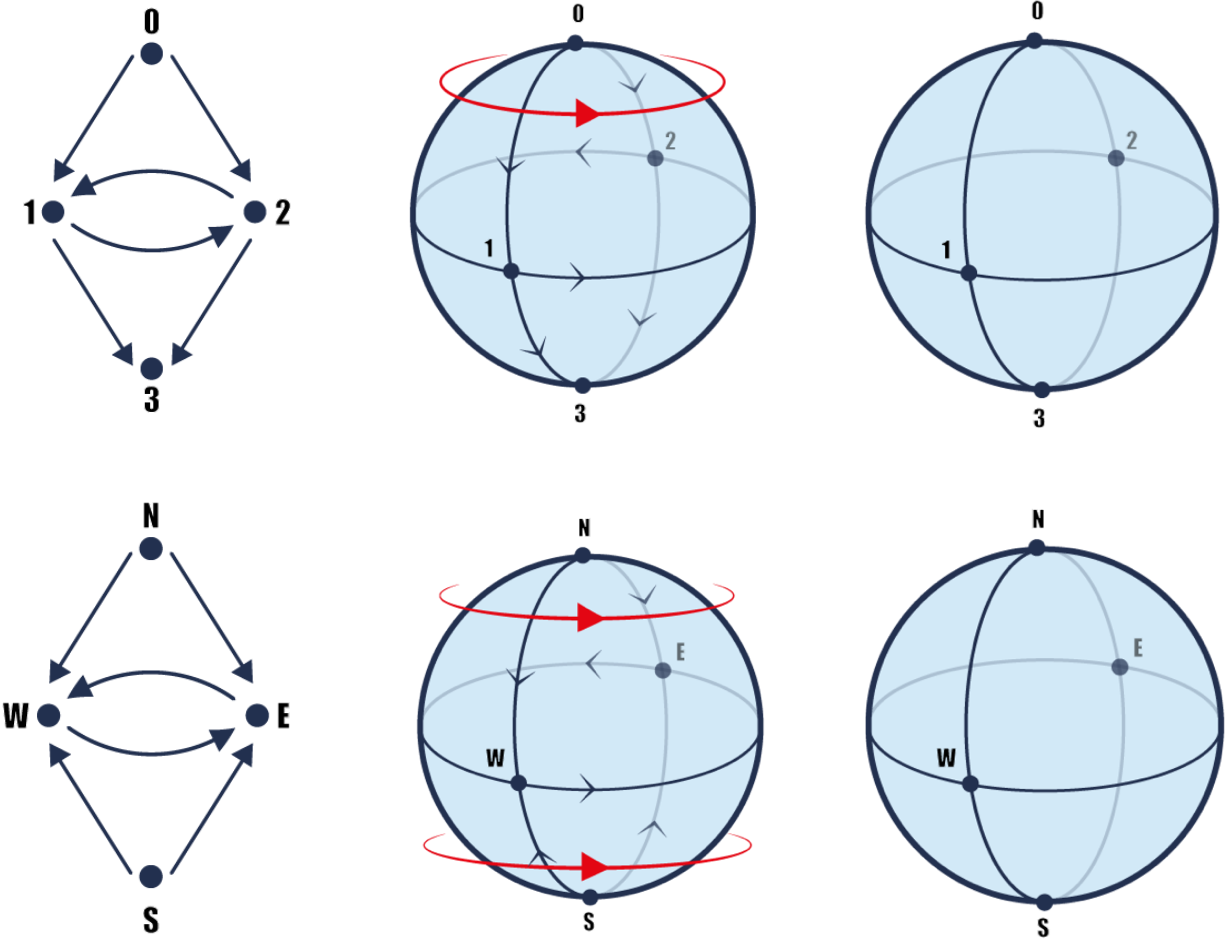} % Adjust the width as needed 
    \caption{ (Left) A pair of non-isomorphic (distinguishable) digraphs, (Middle) their corresponding two-dimensional directed flag complexes, and (Right) their symmetrized undirected versions. Symmetrizing the directed flag complexes collapses the non-isomorphic digraphs into isomorphic \manu{semi-simplicial set}, while the directed flag complexes are non-isomorphic.} %However, examining second-order directed simplicial paths along face maps $(d_1, d_2)$ reveals critical differences in circular flows: the first complex exhibits a flow only in the upper hemisphere, while the second complex shows flows in both the upper and lower hemispheres. This highlights the significance of higher-order directional motifs for distinguishing structural properties. Symmetrizing the directed flag complexes collapses the non-isomorphic digraphs into identical simplicial complexes, illustrating the loss of information through symmetrization, as explained in Remark \ref{rmk:symmetrization}.} 
    \label{fig:circular_flows}
\end{figure}

\textbf{Motivation.} A directed simplicial complex can be transformed into an undirected \manu{semi-simplicial set} via symmetrization, wherein the order of the vertices in each directed simplex is disregarded. The symmetrization preserves isomorphisms, meaning that two isomorphic directed simplicial complexes will remain isomorphic as undirected \manu{semi-simplicial sets} after symmetrization. Informally, this means that two indistinguishable directed simplicial complexes remain indistinguishable after the symmetrization. However, the process is not injective: non-isomorphic directed simplicial complexes may be mapped to isomorphic undirected \manu{semi-simplicial sets}. Informally, this implies that two distinguishable directed simplicial complexes can become indistinguishable after the symmetrization.
For the same reason, composing symmetrization with lifting into (directed) flag complexes -- whether by first symmetrizing a digraph and then lifting it into a flag complex, or by first lifting it into a directed flag complex and then symmetrizing it -- can collapse distinct digraphs into the same undirected \manu{semi-simplicial set}. This is just one of the possible formal hints showing that transitioning to higher-order undirected complexes is not always inherently beneficial, calling for a notion of higher-order directionality. Fig. \ref{fig:circular_flows} illustrates this fact.

\begin{figure}[t]
    \centering
    \includegraphics[width=0.4\columnwidth]{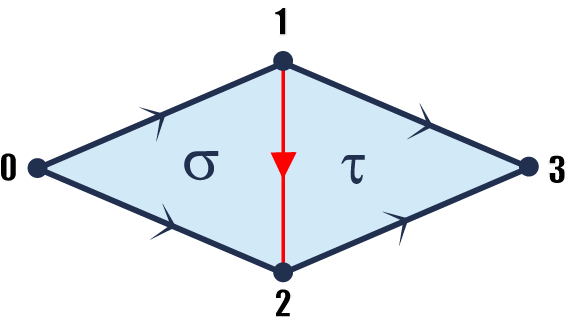} % Adjust the width as needed
    \caption{Example of $(1,0,2)$-adjacent directed 2-simplices $\sigma = (0,1,2)$ and $\tau = (1,2,3)$, sharing the edge $\kappa = (1,2)$ (in red).}
    \label{fig:ij}
\end{figure}

\textbf{Higher-order Topological Directionality.}\label{subsec:dir} We define directed relations among simplices in a directed simplicial complex $\mathcal{K}$ using face maps. Consider a \manu{pair of simplices} $(\sigma, \tau)$ with $\dim(\sigma)=\dim(\tau)$. Let $(d_i, d_j)$ denote a pair of the $i$-th and $j$-th face maps, as defined in \eqref{eq:face_map}. We define $(\sigma, \tau)$ as being \textit{down $(k,i,j)$-adjacent} if there exists a simplex $\kappa$ such that $\dim(\kappa) = \dim(\sigma) - k$ and $d_i(\sigma) \supseteq \kappa \subseteq d_j(\tau)$. To illustrate this, let $\sigma = (0,1,2)$ and $\tau = (1,2,3)$ be 2-simplices (triangles) as in Fig. \ref{fig:ij}. Consider the pair of face maps $(d_0,d_2)$. Applying these maps, we obtain the directed edges $d_0((0,1,2)) = (1,2) = \kappa$ and $d_2((1,2,3)) = (1,2) = \kappa$, making the pair of simplices $(\sigma,\tau)$ down $(1,0,2)$-adjacent. Similarly, we define $(\sigma, \tau)$ as being \textit{up $(k,i,j)$-adjacent} if there exists a simplex $\kappa$ such that $\dim(\kappa) = \dim(\sigma) + k$ and $\sigma \subseteq d_i(\kappa)  \text{ and } \tau \subseteq d_j(\kappa)$.  Notably, for $i \neq j$, if a pair $(\sigma,\tau)$ is up/down $(k,i,j)$-adjacent, then $(\tau, \sigma)$ is up/down $(k,j,i)$-adjacent. Additionally, due to the symmetry in the face maps, if $(\sigma,\tau)$ is up/down $(k,i,i)$-adjacent, then $(\tau, \sigma)$ is also up/down $(k,i,i)$-adjacent, showing that not all the directed adjacencies are necessarily asymmetric. It is then natural to define notions of neighborhood among simplices of the same dimensions using the above adjacencies.

\begin{figure}[t]
    \centering
    \includegraphics[width=\columnwidth]{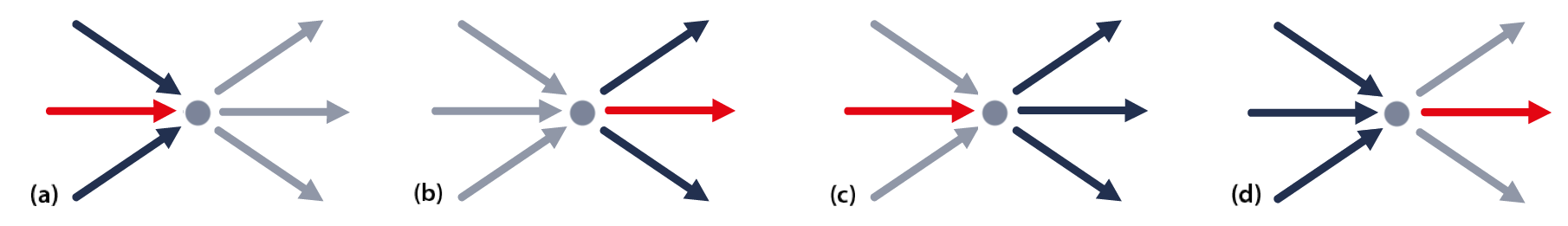} % Adjust the width as needed
    \caption{Examples of edge down adjacencies.  $\sigma$ is the red edge in each subfigure. (a)/(b) $\mathcal{A}_{\downarrow,1}^{0,0}$/$\mathcal{A}_{\downarrow,1}^{1,1}$ connects $\sigma$ with the edges with which it shares a target/source node; (c)/(d) $\mathcal{A}_{\downarrow,1}^{0,1}$/$\mathcal{A}_{\downarrow,1}^{1,0}$ connects $\sigma$ with the edges whose source/target node is the target/source node of $\sigma$.}
    \label{fig:edges}
\end{figure}

\begin{figure}[t]
    \centering
    \includegraphics[width=.7\columnwidth]{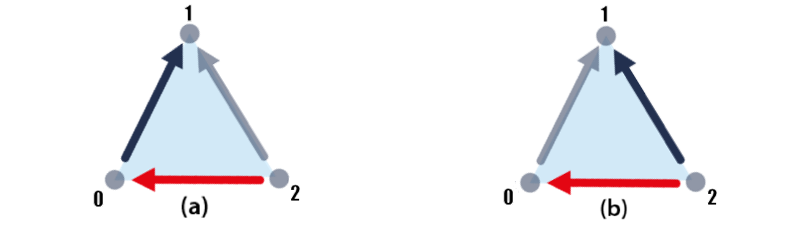} % Adjust the width as needed
    \caption{Examples of edges upper adjacencies. $\sigma$ is the red edge in each subfigure. (a) $\mathcal{A}_{\uparrow,1}^{2,0}$ connects $\sigma$ with the edge on its left; (b) $\mathcal{A}_{\uparrow,1}^{2,1}$ connects $\sigma$ with the edge on its right.}
    \label{fig:triangles}
\end{figure}

 We define the \textit{lower $(k,i,j)$-adjacency} $\mathcal{A}^{ij}_{\downarrow,k}$ of  a simplex $\sigma$ as
\begin{equation}\label{eq:low_adj}
\mathcal{A}^{ij}_{\downarrow,k}(\sigma) =\left\{ 
\hspace{-.2cm}\begin{array}{l}
\displaystyle \tau \in \Sigma |
\end{array}
\hspace{-.3cm}\begin{array}{l}
\, \dim(\sigma) = \dim(\tau),\\
\, \exists \, \manu{\kappa \in \Sigma} \, \text{:}  \dim(\kappa) = \dim(\sigma)- k, \\
\, d_i(\sigma) \supseteq \kappa \subseteq d_j(\tau)
\end{array}
\hspace{-.2cm}\right\}.
\end{equation}
We  define the \textit{upper $(k,i,j)$-adjacency} $\mathcal{A}^{ij}_{\uparrow,k}$ of a simplex $\sigma$ as
\begin{equation}\label{eq:uo_adj}
\mathcal{A}^{ij}_{\uparrow,k}(\sigma) = \left\{
\hspace{-.2cm}\begin{array}{l}
\displaystyle \tau \in \Sigma |
\end{array}
\hspace{-.3cm}\begin{array}{l}
\, \dim(\sigma) =  \dim(\tau), \\
\, \exists \,\manu{\kappa \in \Sigma} \, \text{:}  \dim(\kappa) = \dim(\sigma) + k, \\
\, \sigma \subseteq d_i(\kappa)  \text{ and } \tau \subseteq d_j(\kappa) 
\end{array}
\hspace{-.2cm}\right\}.
\end{equation}
For the lower  $(k,i,j)$-adjacency in \eqref{eq:low_adj} we assume $\dim(\kappa)=0$ if $k>\dim(\sigma)$, while for the upper  $(k,i,j)$-adjacency in \eqref{eq:uo_adj} we assume $\dim(\kappa)=\dim(\mathcal{K})$ if $k>\dim(\mathcal{K})-\dim(\sigma)$. We show some examples of edge lower and upper adjacencies in Fig. \ref{fig:edges} and Fig. \ref{fig:triangles}, respectively. The upper $(k,i,j)$-adjacency offers a complementary perspective to the lower $(k,i,j)$-adjacency, because the former captures the directed interactions where simplices $\sigma$ and $\tau$ are both included in some higher-order simplices, while the latter when they both contain some lower-order simplices. Finally, face maps can also be used to define notions of neighborhood among simplices of different dimensions. %Let $\sigma \in \mathcal{K}$ be an $n$-simplex. The $i$-th face map $d_i$ for $0 \leq i \leq n$ yields the $i$-th face of $\sigma$. The boundary map $\partial$ of $\sigma$ is given by the alternating sum of its $i$-th faces: $\partial(\sigma) = \sum_{i=0}^{n} (-1)^i d_i(\sigma)$, where $d_i(\sigma)$ denotes the $i$-th face of $\sigma$ \cite{hatcher2005algebraic}. The preimage (or fiber) of $d_i$, denoted $d_i^{-1}(\sigma)$, consists of all $(n+1)$-simplices that contain $\sigma$ as their $i$-th face. With a slight abuse of terminology, we refer to $\mathcal{B}^i(\sigma) = d_i(\sigma)$ as the \textit{$i$-th boundary} of $\sigma$, and $\mathcal{C}^i(\sigma) = d_i^{-1}(\sigma)$ as the \textit{$i$-th coboundary set}. Accordingly, we define the \textit{boundary} and \textit{coboundary} sets as: 
%\begin{align}
    %&\mathcal{B}(\sigma) = \{\tau \in \Sigma| \tau \in \{d_i(\sigma)\}_{i = 0}^{\dim(\sigma)}\}, \\
    %&\mathcal{C}(\sigma) = \{\tau \in \Sigma| \tau \in \bigcup_id^{-1}_{i=0}^{\dim(\sigma)+1}(\sigma)\},
%\end{align} 
We define the \textit{boundary} $\mathcal{B}$ and \textit{coboundary} $\mathcal{C}$ of $\sigma \in \mathcal{K}$ as
\begin{align}&\mathcal{B}(\sigma) = \bigcup_{i = 0}^{\dim(\sigma)} \{d_i(\sigma)\},  &\mathcal{C}(\sigma) = \bigcup_{i = 0}^{\dim(\sigma)+1} d_i^{-1}(\sigma),\end{align}
where $d^{-1}_i$ is the preimage of $d_i$. The boundary and the coboundary of $\sigma$ are then its facets and the simplices it is a facet of, respectively.  

\begin{figure}[t]
    \centering
    \includegraphics[width=0.55\columnwidth]{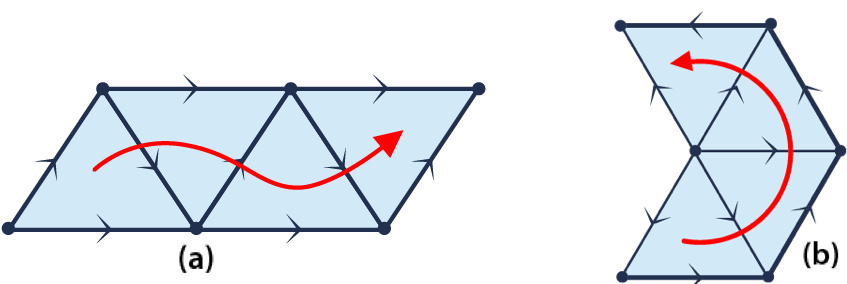} % Adjust the width as needed
    \caption{Examples of simplicial paths (in red) of 2-simplices (triangles). (a) The $(1, 0, 2)$ path, showing the simplices are equidirected; (b) the $(1, 1, 2)$ path, revealing a circular flow around a source node.}
    %\caption{Both complexes admit multiple directed paths at the graph level; however, a unique simplicial path along the triangles reveals an inherent high-order directionality. The depicted paths (in red) are: (a) the $(1,0,2)$ simplicial path, and (b) the $(1,1,2)$ simplicial path.}
    %\caption{Both complexes admit multiple directed paths at the graph level; however, only a unique simplicial path exists along the triangles, revealing an inherent second-order directional structure. The depicted paths (in red) are: (a) the $(1,0,2)$ simplicial path, and (b) the $(1,1,2)$ simplicial path, illustrating a circular flow around a source node.}
    %\caption{Both complexes admit multiple directed paths at the graph level, but only a unique simplicial path exists along the triangles, revealing an inherent second-order directional structure. The depicted paths (in red) are: (a) the $(1,0,2)$ simplicial path and (b) the $(1,1,2)$ simplicial path, illustrating a circular flow around a source node.}
    \label{fig:simplices}
\end{figure}

\textbf{Consistency of Higher-order Topological Directionality.} The way we define the directed adjacencies among simplices is grounded in the notion of simplicial directed paths \cite{Riihimaki24qconnect}. In a digraph $G = (V,E)$, a directed path is defined as a sequence of vertices $ (v_0, v_1, \dots, v_n) $ where each consecutive pair $(v_i, v_{i+1}) \in E$ forms a directed edge. In a directed simplicial complex $\mathcal{K}$, a $(k,i,j)$-simplicial path between an pair of simplices $(\sigma, \tau)$ in $\mathcal{K}$ is a sequence of simplices $\sigma = \alpha_0, \alpha_1, \alpha_2, \dots, \alpha_n, \alpha_{n+1} = \tau$ such that each consecutive pair $(\alpha_k, \alpha_{k+1})$ is $(k,i,j)$-adjacent along the face maps $(d_i, d_j)$. We show some examples of simplicial paths of triangles in Fig. \ref{fig:simplices}. Higher-order directionality reveals novel, discriminative structural properties, as demonstrated again in Fig. \ref{fig:circular_flows}, where circular flows (in red) emerge considering $(1,0,2)$-simplicial paths. Finally, simplicial paths can also traverse simplices of different orders. We decided to add the constraint $\dim(\sigma)=\dim(\tau)$ in \eqref{eq:low_adj}-\eqref{eq:uo_adj} to keep a distinction between same-dimension and different-dimension neighbors.

\begin{figure*}[ht!]
  \centering
  \hspace{.5cm}\begin{subfigure}{.4\linewidth}
  \centering
\includegraphics[width=\linewidth]{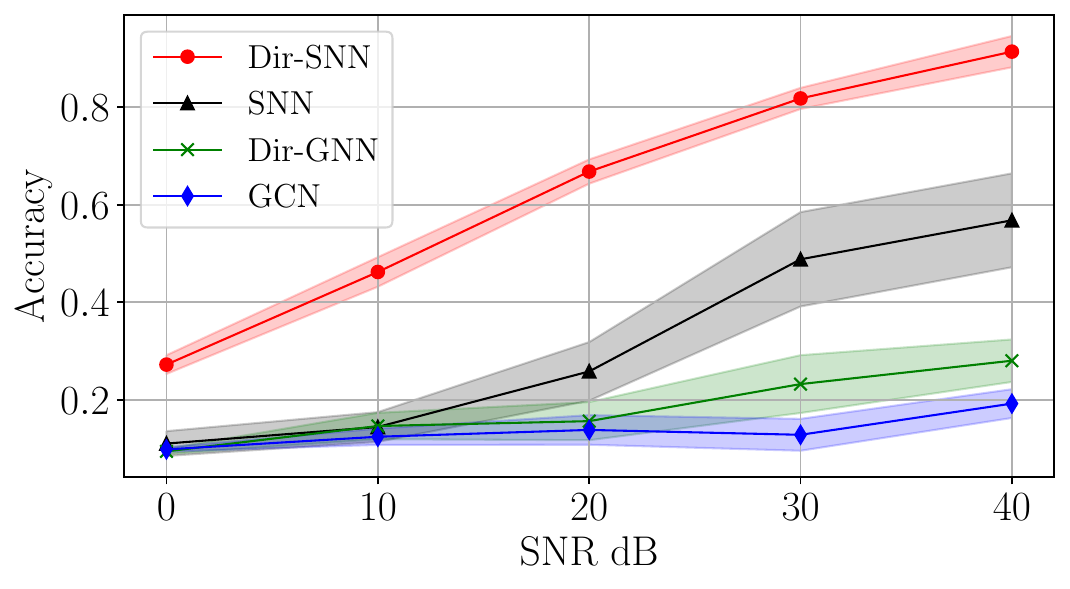}
  \end{subfigure}
  \hfill
   \begin{subfigure}{.4\linewidth}     
    \centering
\hspace{-2cm}\includegraphics[width=\linewidth]{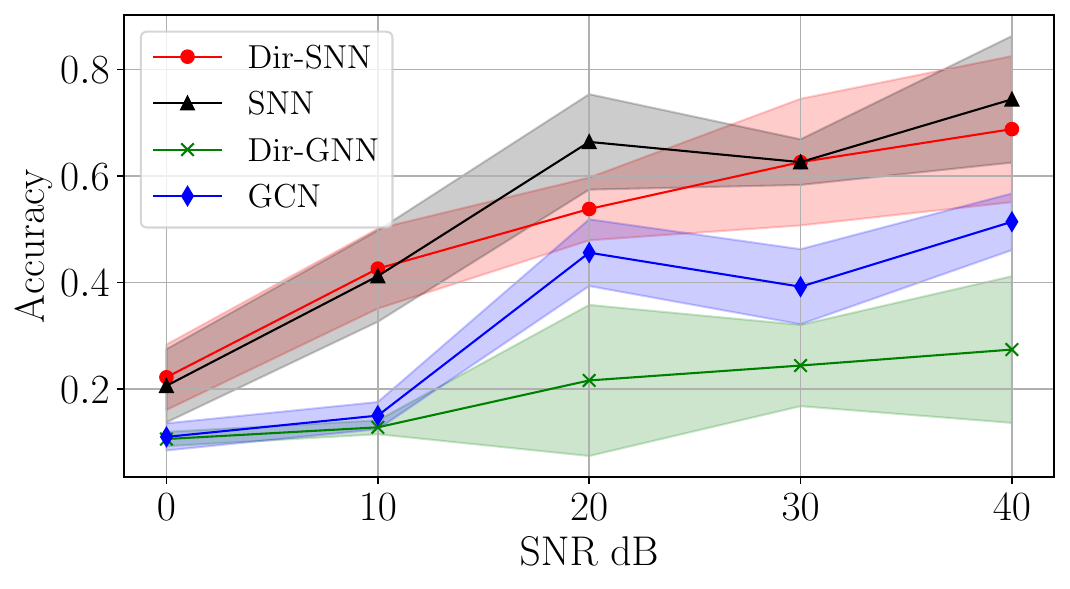}
  \end{subfigure}
    \caption{SNR vs accuracy of directed and undirected TNNs and GNNs on directed (left) and undirected (right) synthetic flag complexes.}
  \label{fig:dirtnn_tnn_snr}
\end{figure*}

\textbf{Directed Simplicial Neural Networks} Directed Simplicial Neural Networks (Dir-SNNs) are message-passing networks \cite{gilmer2017} leveraging the adjacencies from \eqref{eq:low_adj}-\eqref{eq:uo_adj}.  Given (i) a directed simplicial complex $\mathcal{K}$, and (ii) sets $\{\mathcal{A}^{ij}_{\downarrow,k}\}_{i,j,k}$ and $\{\mathcal{A}^{ij}_{\uparrow,k}\}_{i,j,k}$ collecting \textit{some} of the lower and upper $(k,i,j)$-adjacencies as in \eqref{eq:low_adj}-\eqref{eq:uo_adj}, respectively, the $l$-th layer of a Dir-SNN updates the feature vector $\mathbf{x}^l_\sigma$ of a  $\sigma \in \mathcal{K}$ as
\begin{align}
    &\mathbf{m}^{l+1}_{\sigma,\downarrow^{ijk}} =  \bigoplus_{\tau \in \mathcal{A}^{ij}_{\downarrow,k}(\sigma)} \psi_{\mathcal{A}^{ij}_{\downarrow,k}}\left(\mathbf{x}^l_\sigma,\mathbf{x}^l_\tau,\mathbf{x}^l_\kappa\right), \label{eq:message_low}\\
    &\mathbf{m}^{l+1}_{\sigma,\uparrow^{ijk}} =  \bigoplus_{\tau \in \mathcal{A}^{ij}_{\uparrow,k}(\sigma)} \psi_{\mathcal{A}^{ij}_{\uparrow,k}}\left(\mathbf{x}^l_\sigma,\mathbf{x}^l_\tau,\mathbf{x}^l_\kappa\right), \label{eq:message_up}\\
    &\mathbf{m}^{l+1}_{\sigma,\mathcal{B}} = \bigoplus_{\tau \in \mathcal{B}(\sigma)}\psi_{\mathcal{B}}\left(\mathbf{x}^l_\sigma,\mathbf{x}^l_{\tau}\right),\label{eq:message_boundary} \\
    &\mathbf{m}^{l+1}_{\sigma,\mathcal{C}} = \bigoplus_{\tau \in \mathcal{C}(\sigma)}  \psi_{\mathcal{C}}\left(\mathbf{x}^l_\sigma,\mathbf{x}^l_\tau\right),\label{eq:message_coboundary} \\
    &\mathbf{x}^{l+1}_\sigma = \phi(\mathbf{x}^l_\sigma, \{\mathbf{m}^{l+1}_{\sigma,\downarrow^{ijk}}\}_{ijk}, \{\mathbf{m}^{l+1}_{\sigma,\uparrow^{ijk}}\}_{ijk}, \mathbf{m}^{l+1}_{\sigma,\mathcal{B}},\mathbf{m}^{l+1}_{\sigma,\mathcal{C}}). 
\end{align}

with $\kappa$ in \eqref{eq:message_low} as in \eqref{eq:low_adj} and in \eqref{eq:message_up} as in \eqref{eq:uo_adj}, $\bigoplus$ being a intra-neighborhood aggregator. The neighborhood-dependent message functions $\psi_{\mathcal{A}^{ij}_{\downarrow,k}}$, $\psi_{\mathcal{A}^{ij}_{\uparrow,k}}$, $\psi_{\mathcal{B}}$ and $\psi_{\mathcal{C}}$, and the update function $\phi$ are learnable functions. In other words, the feature vector of a simplex is updated in a learnable fashion through aggregated messages with its neighboring simplices.  At the $l$-th layer, a simplex has collected information from simplices that are up to $l$ steps away from it along the simplicial directed paths induced by the chosen adjacencies. 

\textbf{\textit{Remark.}} In \eqref{eq:message_boundary} and \eqref{eq:message_coboundary}, single message functions $\psi_{\mathcal{B}}$ and $\psi_{\mathcal{C}}$ are used for computational efficiency. However, face map- and preimage-dependent message functions, i.e. $\psi_{d_i}$ and $\psi_{d_i^{-1}}$, could leverage more fine-grained directed information, e.g. edges communicating with their source or target nodes using different sets of weights.

\textbf{Expressiveness of Dir-SNNs.} The expressive power of topological neural networks (including GNNs) is usually measured by their capacity to distinguish non-isomorphic objects within their underlying domain \cite{xu2018powerful}.  Here, we introduce the following result.

\textbf{\textit{Theorem 1.}} There exist Dir-SNNs that are more powerful than Directed GNNs (Dir-GNNs) \cite{Rossi23} at distinguishing non-isomorphic digraphs using a directed flag complex lifting.

\textbf{\textit{Proof.}} See Appendix.%.

An example is shown in Fig. \ref{fig:dswl}. Intuitively, our proof follows the same strategy of \cite{bodnar2021weisfeiler}, thus it relies on the definition of a directed simplicial isomorphism test that is proven to be (i) an upper bound on the expressiveness of Dir-SNNs, and (ii) more powerful than the directed graph isomorphism test \cite{grohe2021colordirwl}, being an upper bound on the expressiveness of Dir-GNNs \cite{Rossi23}. 

\textit{\textbf{Remark.}} Dir-SNNs generalize Dir-GNNs \cite{Rossi23}. A Dir-GNN is a Dir-SNNs operating on a digraph, i.e., a directed simplicial complex of dimension one, which updates the node feature vectors using only the $(1,0,1)$- and the $(1,1,0)$-upper adjacencies in \eqref{eq:uo_adj}.  %Dir-SNNs do not generalize (as they operate on different objects) but are related to message-passing networks on undirected simplicial complexes \cite{bodnar2021weisfeiler}. In particular, given an undirected simplicial complex and any reference order to make it directed, the lower and upper adjacencies \cite{bodnar2021weisfeiler} are recovered by taking the union of the directed adjacencies obtained setting $k=1$ and using all the possible $i,j$ in \eqref{eq:low_adj} and all the possible distinct $i,j$ in \eqref{eq:uo_adj}. Again, the standard boundary and coboundary are obtained using $\mathcal{I} = \{0,1,\dots,\dim(\mathcal{K})\}$ as incidence set. 

% REMEMBER: mention that flag complexes, besides being isomorphism preserving are also injective constructions on non-isomorphic graphs.
\begin{figure}[t]
    \centering
    \includegraphics[width=0.85\columnwidth]{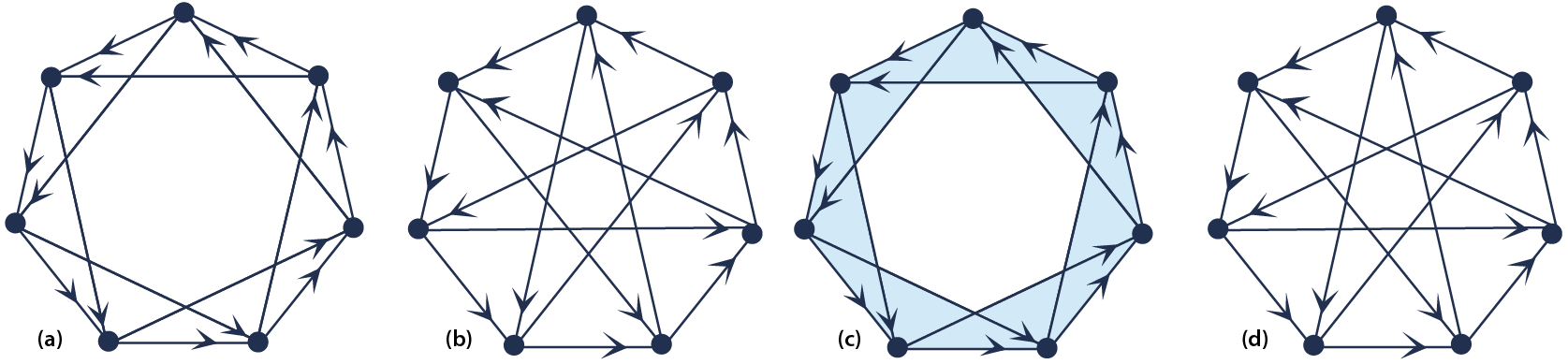} % Adjust the width as needed
    \caption{A pair of non-isomorphic directed graphs (a) and (b), along with their corresponding flag complexes (c) and (d). These digraphs can be distinguished by Dir-SNNs but not by Dir-GNNs \cite{Rossi23}.}
    \label{fig:dswl}
\end{figure}
\vspace{-.2cm}
\section{Numerical Results}
\vspace{-.1cm}
\label{sec:anr}

We first provide some preliminary results to validate the effectiveness of Dir-SNN on a synthetic source localization task at the edge level \cite{fiorellino2024tnn}. Then, we perform a simple experiment to validate Theorem 1.\footnote{Code at \url{https://github.com/ManuelLecha/DirSNN}}
\vspace{-.6cm}
\subsection{Source Localization}\label{sec:source_loc}
\vspace{-.1cm}
 \textbf{Dataset.} 
We generate directed and undirected graphs following a Stochastic Block Model~\cite{holland1983sbm}. Each graph has 70 nodes uniformly divided into 10 communities, with intra- and inter-community edge probabilities of 0.9 and 0.01, respectively. Intra-community edges are grouped into 10 edge communities, and the remaining inter-community edges form an 11th partition. 
We generate 1000 edge signals from a zero-mean Gaussian distribution with variance $1/N_\textnormal{edges}$, and then, for each signal, we introduce spikes to randomly selected source edges belonging to a single community with intensity $\alpha \sim \mathcal{N}(0, 1)$. In the directed case, the graphs are lifted in their corresponding directed flag complexes, and the spikes are diffused over the graph following $\textbf{x}' = \textbf{S}^t \textbf{x} + \textbf{n}$, where $\textbf{S}\in \mathbb{R}^{N_\textnormal{edges} \times N_\textnormal{edges}}$ is a non-symmetric binary matrix encoding $\mathcal{A}_{\downarrow,1}^{0,1}$ from \eqref{eq:low_adj} for each edge, $t$ is the order of diffusion sampled from a Student-T distribution with 10 degrees of freedom and capped at 100, $\textbf{x}$ is the original signal with the added spikes and $\textbf{n}$ is additive white Gaussian noise inducing a specific SNR. 
In the undirected case, the graphs are lifted in their corresponding flag complexes, and $\textbf{S}$ is the symmetric lower edge adjacency, analogously to~\cite{fiorellino2024tnn}.
The task is to identify the community originating the spikes, thus a classification problem with 11 classes. 

\textbf{Experimental setup.} 
Since the designed task considers only edge signals, we decided to employ a specific instance of Dir-SNN that operates on the four directed adjacency relations described by $\mathcal{A}_{\downarrow,1}^{0,0}$, $\mathcal{A}_{\downarrow,1}^{0,1}$, $\mathcal{A}_{\downarrow,1}^{1,0}$, $\mathcal{A}_{\downarrow,1}^{1,1}$, without considering boundary and coboundary. The aggregators, update, and message functions are chosen to have a convolutional architecture \cite{kipf2016semi}.  Therefore, we compare with an undirected convolutional SNN using the undirected lower adjacency of the edges~\cite{yang2023convolutional}, and with two GNNs that operate on the projection of edge features on their endpoint nodes, one directed (Dir-GNN~\cite{Rossi23}) and the other undirected (GCN~\cite{kipf2016semi}).
We perform a grid search for all the models among the following values: $\{1,2,3\}$ layers of size $\{16,32,64\}$.  
The embeddings are max-pooled and fed into an MLP to perform classification. We average over 5 seeds. 

 \textbf{Discussion.} In Fig.~\ref{fig:dirtnn_tnn_snr}, Dir-SNN consistently and largely outperforms all the baselines for all levels of SNR on directed flag complexes (Left). 
Moreover,  Dir-SNN performs comparably to SNN on flag complexes (Right), showing their robustness to model mismatching.
\vspace{-.25cm}
\subsection{Expressivity Validation}
\vspace{-.1cm}
 \textbf{Dataset.} We employ a toy dataset containing only the two graphs in Fig. \ref{fig:dswl} (a)-(b), and we assign them two different classes. The task is then to correctly classify them.

\textbf{Experimental Setup.} We compare Dir-SNN and Dir-GNN. The parameters are the same of \ref{sec:source_loc}, but for Dir-SNN we also use $\mathcal{A}_{\uparrow,1}^{2,0}$ (to leverage the directed triangles). The two graphs in Fig. \ref{fig:dswl} (a)-(b) are fed to the Dir-GNN with constant features on each node, while the two corresponding directed flag complexes in Fig. \ref{fig:dswl} (c)-(d) are fed to the Dir-SNN with constant features on each simplex.
\begin{table}[t]
\centering
\vspace{.5cm}
\begin{tabular}{|c|c|}
\hline
\textbf{Model} & \textbf{Accuracy} \\ \hline
Dir-GNN        & \textcolor{Mahogany}{50\%}              \\ \hline
Dir-SNN        & \textcolor{ForestGreen}{100\%}             \\ \hline
\end{tabular}
\caption{Discrimination accuracies of Dir-GNN and Dir-SNN.}\label{table:expr}
\end{table}

\textbf{Discussion.} The results in Table \ref{table:expr} numerically validate the improved expressivity of Dir-SNN stated in Theorem 1. Indeed, Dir-GNN cannot discriminate the two graphs, while Dir-SNN can discriminate them thanks to the lifting into directed flag complexes.

\vspace{-.25cm}
\section{Conclusions}
\label{sec:conclusions}
\vspace{-.1cm}
We introduced \textit{Directed Simplicial Networks} (Dir-SNNs), the first family of message-passing networks operating on directed simplicial complexes and leveraging \textit{novel notions of higher-order topological directionality}. Dir-SNNs effectively model complex directed and possibly asymmetric relationships that are inaccessible to traditional directed graph-based or undirected topological models, showing improved expressivity. Numerical experiments validated the effectiveness and the expressivity of Dir-SNNs. In the journal version of this paper, we will characterize the expressivity of Dir-SNNs more comprehensively. Moreover, we will develop a spectral theory to better characterize the (implicit, at the moment) filtering operations in Dir-SNNs. Finally, we will exhaustively evaluate Dir-SNNs on a wide range of real data, both on graph and higher-order tasks. 

\newpage
\bibliographystyle{ieeetr}
\bibliography{refs}
%\end{document}

\appendix
\label{app}

The expressive power of Topological Neural Networks (TNNs) is usually (but not only \cite{zhang2024beyond}) measured by evaluating their ability to distinguish isomorphic combinatorial topological spaces (CTS) \cite{bodnar2021weisfeiler}. The Weisfeiler-Leman (WL) test, an isomorphism test for graphs \cite{weisfeiler1968reduction}, has been proven to be an upper bound on GNNs discriminative power \cite{xu2018powerful}, i.e. no vanilla GNN can distinguish isomorphic graphs better than the WL. Graph liftings, such as the directed flag complex lifting (introduced in Sec. \ref{sec:background}), canonically map graphs to Combinatorial Topological Spaces (CTS), enriching their structure with higher-order relations while preserving the original graph’s nodes (0-simplices) and edges (1-simplices). These liftings ensure that isomorphic graphs are mapped to isomorphic CTSs and non-isomorphic graphs to non-isomorphic CTSs. For this reason, topological variants of the WL test have been developed to exploit the additional higher-order structure introduced by graph lifting \cite{bodnar2021weisfeiler, bodnarcwnet, truong2024wlpath}. These extensions, as their graph counterpart, have been proven to be an upper bound on TNNs ability to distinguish isomorphic CTS, and, more importantly, they have been used to prove that TNNs operating on lifted graphs are more expressive than traditional GNNs. In other words, TNNs have demonstrated a greater capacity to distinguish between non-isomorphic graphs that GNNs may otherwise classify as identical. In this appendix, we prove Theorem 1, i.e. we formally show that Directed Simplicial Neural Networks (Dir-SNNs) leveraging higher-order structures and directional information through directed flag complex liftings are strictly more expressive than Directed Graph Neural Networks (Dir-GNNs) \cite{Rossi23}.  Our proof is a non-trivial adaptation of the  Simplicial Weisfeiler-Leman (SWL) test from \cite{bodnar2021weisfeiler} to the directed setting (defined in Sec. \ref{sec:spnns}). In particular, we first introduce the Directed Simplicial Weisfeiler-Leman (D-SWL) test, the first isomorphism test for directed simplicial complexes. We then rigorously show that D-SWL is a more powerful isomorphism test than D-WL \cite{grohe2021colordirwl}, being an upper bound on the expressiveness of Dir-GNNs. Finally, the proof is concluded by showing that the D-SWL is an upper bound on the expressiveness of Dir-SNNs. The structure of our proofs, theorems, and notation follows closely the work of \cite{bodnar2021weisfeiler}, and we encourage readers to consult it for further context and foundational concepts.

\textbf{Directed Simplicial Weisfeiler-Leman (D-SWL) Test.} We assume k = 1  and omit this specification from the notation of the up- and down-adjacencies in \eqref{eq:uo_adj}-\eqref{eq:low_adj}. Let $\mathcal{K}$ be a directed simplicial complex, and let $c^t$ denote the colouring of simplices at iteration $t$ in the D-SWL test. Let $\sigma \in \mathcal{K}$ be an $n$-simplex, we define the following arguments at step $t$:

\textit{1. Boundary colouring:} Represents the ordered tuple of colours assigned to each element of the boundary of $\sigma$, $d_i(\sigma)$. \[c_{\mathcal{B}}^t(\sigma) = (c_{d_0(\sigma)}, \dots, c_{d_n(\sigma)}).\]

\textit{2. Coboundary colouring:} Let $c_{d_i^{-1}(\sigma)} = \{\{c_{\tau} \mid \tau \in d_i^{-1}(\sigma)\}\}$ be the multiset of colours for the preimage of the $i$-th face map of $\sigma$. Then the coboundary colouring of $\sigma$ is given by:
\[
c_{\mathcal{C}}^t(\sigma) = \bigcup_{i=0}^{\operatorname{dim}(\sigma) + 1} c_{d_i^{-1}(\sigma).}
\]

\textit{3. Down adjacency colouring:} Captures the pairs of colours associated with simplices $\tau$ that are downward adjacent to $\sigma$, where $\kappa$ represents a shared face. \[(c_{\downarrow}^{ij})^t(\sigma) = \{\{(c_{\tau}^t, c_\kappa) \mid \tau \in \mathcal{A}^{ij}_\downarrow(\sigma), d_i(\sigma) = \kappa = d_j(\tau)\}\}.\]

\textit{4. Up adjacency colouring:} Represents the pairs of colours of simplices $\tau$ that are upward adjacent to $\sigma$, where $\kappa$ represents a shared higher-dimensional face. \[(c_{\uparrow}^{ij})^t(\sigma) = \{\{(c_{\tau}^t, c_\kappa) \mid \tau \in \mathcal{A}^{ij}_\uparrow(\sigma), d_i(\kappa) = \sigma, d_j(\kappa) = \tau\}\}.\]

The D-SWL test distinguishes non-isomorphic directed simplicial complexes through iterative color refinement of simplices. The procedure consists of the following steps:

\textit{1) Initialization:} At  $t = 0$, assign an initial colouring to all simplices.

\textit{2) Color Propagation and Refinement:} For each simplex  $\sigma$, propagate and refine the colours using the injective update rule:
\[
c^{t+1}_\sigma = \operatorname{HASH}\left(c_\sigma^t, c_{\mathcal{B}}^t(\sigma), c_{\mathcal{C}}^t(\sigma), ((c_{\downarrow}^{ij})^t(\sigma))_{ij}, ((c_{\uparrow}^{ij})^t(\sigma))_{ij}\right),
\]
where \( i, j \in \{0, \dots, \dim(\sigma)\} \) and \( i, j \in \{0, \dots, \dim(\sigma)\} \) for the down and upper adjacencies, respectively.

\textit{3) Termination: }Repeat this process until the colouring stabilizes. Two directed simplicial complexes are considered non-isomorphic if their stable color histograms differ.
 
We now demonstrate that, for distinguishing non-isomorphic directed simplicial complexes, the multisets of colours associated with the coboundaries and down-adjacencies can be removed without sacrificing the expressiveness of the D-SWL test. This will simplify the rest of the proof, and it also demonstrates that D-SWL (and, eventually, Dir-SNNs) can be made computationally more efficient.

\begin{lemma}\label{lemma:dwslatleast}
D-SWL with $\operatorname{HASH}\left(c_\sigma^t,  c_{\mathcal{B}}^t(\sigma), ((c_{\uparrow}^{ij})^t(\sigma))_{i,j}\right)$ is as powerful as D-SWL with the generalised update rule $c^{t+1}_\sigma = \operatorname{HASH}\left(c_\sigma^t, c_{\mathcal{B}}^t(\sigma), c_{\mathcal{C}}^t(\sigma), ((c_{\downarrow}^{ij})^t(\sigma))_{i,j}, ((c_{\uparrow}^{ij})^t(\sigma))_{ij}\right)$.
\end{lemma}

\begin{proof}First, we show that coboundary multisets can be removed, following Lemma 25 of \cite{bodnar2021weisfeiler}.\subsection{Omitting $c_{\mathcal{C}}^t(\sigma)$}
%We begin by proving that the multiset of coboundary colours $c_{\mathcal{C}}^t(\sigma)$ can be removed from the generalized update rule of D-SWL without sacrificing expressiveness. 
Let $a^t$ and $b^t$ denote the colourings at iteration $t$ for the generalized update rule (which includes the coboundary color multisets) and the restricted update rule (which excludes them), respectively. To establish that these two colourings are equivalent, we need to show that $a^t$ refines $b^t$ and vice versa. Since the refinement $a^t$ to $b^t$ is straightforward, we focus on proving that $b^t$ refines $a^t$. We proceed by induction on $t$.

\subsubsection{Base Case}
At $t = 0$, all simplices are assigned the same initial color, so the base case holds trivially.

\subsubsection{Inductive Hypothesis}
Assume that for some iteration $t$, the restricted colouring $b^t$ refines the generalized colouring $a^t$. That is, if $b^t_\sigma = b^t_\tau$ for two simplices $\sigma \in K_1$ and $\tau \in K_2$ with $\dim(\sigma) = \dim(\tau)$, then $a^t_\sigma = a^t_\tau$. We aim to show that this holds for the next iteration, $t+1$, i.e., $b^{t+1}_\sigma = b^{t+1}_\tau$ implies $a^{t+1}_\sigma = a^{t+1}_\tau$.

\subsubsection{Inductive Step}
Suppose that $b^{t+1}_\sigma = b^{t+1}_\tau$. This implies equivalence in the arguments of the update rule at time $t$. In particular, the upward adjacency colouring multisets $(b_{\uparrow}^{ij})^t(\sigma)$ include the colours of all higher-dimensional simplices $\delta_\sigma$ whose $i$-th face is $\sigma$. This accounts for the colours of the elements in the preimage of the $i$-th face map $d_i^{-1}(\sigma)$. Thus, we have:\[
b_{d_i^{-1}(\sigma)} = b_{d_i^{-1}(\tau)}
\]
for all $i$. Therefore, by definition, $b_{\mathcal{C}}^t(\sigma) = b_{\mathcal{C}}^t(\tau)$. Finally, by the inductive hypothesis, the arguments for the generalized hash update rule must also be equivalent at time $t$, so $a^{t}_\sigma = a^{t}_\tau$, leading to $a^{t+1}_\sigma = a^{t+1}_\tau$, as required, completing the proof.\\

Second, we prove that the update rule can be further refined by omitting the downward adjacencies.

\subsection{Omitting $((c_{\downarrow}^{ij})^t(\sigma))_{ij}$}

Let $a^t$ and $b^t$ denote the colourings at iteration $t$ for the generalized update rule (which includes down-adjacencies but excludes coboundary color multisets) and the restricted update rule (which excludes both down-adjacencies and coboundary color multisets), respectively. Our goal is to show that $b^{2t}$ refines $a^t$ by induction.

\subsubsection{Base Case}
At $t = 0$, all simplices are assigned the same initial color, so the base case trivially holds.

\subsubsection{Inductive Hypothesis}
Assume that for some iteration $t$, the restricted colouring $b^{2t}$ refines the generalized colouring $a^t$. This means that if $b^{2t}_\sigma = b^{2t}_\tau$ for two simplices $\sigma \in K_1$ and $\tau \in K_2$ with $\dim(\sigma) = \dim(\tau)$, then $a^t_\sigma = a^t_\tau$. We aim to show that this holds for the next iteration, $t+2$.

\subsubsection{Inductive Step}
Suppose that $b^{2t+2}_\sigma = b^{2t+2}_\tau$. This implies that $b^{2t}_\sigma = b^{2t}_\tau$, meaning the arguments for the restricted update rule at step $2t$ are equivalent. We now need to prove that $(b^{ij}_{\downarrow})^{2t}(\sigma) = (b^{ij}_{\downarrow})^{2t}(\tau)$ holds for all pairs $(i,j)$. Assume, by contradiction, that there exists a pair of indices $(i,j)$ such that $(b^{ij}_{\downarrow}(\sigma))^{2t} \neq (b^{ij}_{\downarrow}(\tau))^{2t}$. Without loss of generality, suppose there is a pair of colours $(c_0, c_1)$ that appears more frequently in $(b^{ij}_{\downarrow}(\sigma))^{2t}$ than in $(b^{ij}_{\downarrow}(\tau))^{2t}$. For simplicity, as in \cite{bodnar2021weisfeiler}, assume $b^{2t}_\sigma \neq c_0 \neq b^{2t}_\tau$. Define %such that $b_\delta^{2t} = c_1$, 
$c_\delta^j = b_{d_j^{-1}(\delta)}(c_0)$ to denote the multiplicity of $c_0$ in the set $b_{d_j^{-1}(\delta)}$, i.e., the number of elements in $d_j^{-1}(\delta)$ coloured with $c_0$.  Now, assume there exist two simplices $\delta_1$ and $\delta_2$, such that $c_{\delta_1}^j \neq c_{\delta_2}^j$ (without loss of generality, assume $c_{\delta_1}^j > c_{\delta_2}^j$).  Thus, for all $s \in \{0,\dots,\operatorname{dim}(\delta_1) + 1\}$, $c_0$ appears more frequently in $(b^{js}_{\uparrow})^{2t}(\delta_1)$ than in $(b^{js}_{\uparrow})^{2t}(\delta_2)$. This implies that $b^{2t+1}_{\delta_1} \neq b^{2t+1}_{\delta_2}$. Since by assumption, the number of tuples $(c_0,c_1)$ is greater in $(b^{ij}_{\downarrow}(\sigma))^{2t}$ than in $(b^{ij}_{\downarrow}(\tau))^{2t}$, we can substitute $\delta_1 = d_i(\sigma)$, and $\delta_2 = d_i(\tau)$. This yields $c^j_{d_i(\sigma)} = b_{d_j^{-1}(d_i(\sigma))}(c_0) \neq c^j_{d_i(\tau)} = b_{d_j^{-1}(d_i(\tau))}(c_0)$. Therefore, we conclude: \[
b^{2t+1}_{d_i(\sigma)} \neq b^{2t+1}_{d_i(\tau)},
\] which implies that the boundary colouring tuples $b^{2t+1}_{\mathcal{B}}(\sigma) \neq b^{2t+1}_{\mathcal{B}}(\tau)$. Hence, $b^{2t+2}_\sigma \neq b^{2t+2}_\tau$, leading to a contradiction. Therefore, $(b^{ij}_{\downarrow})^{2t}(\sigma) = (b^{ij}_{\downarrow})^{2t}(\tau)$ for all pairs $(i,j)$. Finally, by the inductive hypothesis, $b^{2t}$ refines $a^t$, completing the proof.
\end{proof}

\textbf{D-SWL vs D-WL.} We now prove the following result.

\begin{lemma}\label{thm:dswl-dwl}
D-SWL with a directed flag complex lifting is strictly more powerful than D-WL.
\end{lemma}
\begin{proof} We begin by proving that D-SWL is at least as expressive as D-WL in distinguishing non-isomorphic directed simplicial complexes. The proof follows closely the structure of Lemma 27 in \cite{bodnar2021weisfeiler}, with key differences in the treatment of upper adjacencies: D-SWL and D-WL employ distinct colouring multisets to capture asymmetric directional dependencies.
\setcounter{subsection}{0}
\subsection{At Least As Expressive}

 Let $\mathcal{K}$ be a directed simplicial complex. Denote by $a^t$ and $b^t$ the colourings of the same vertices in $\mathcal{K}$ at iteration $t$ of D-WL and D-SWL, respectively. We aim to show that the D-SWL colouring $b^t$ refines the D-WL colouring $a^t$.%, i.e., for any two vertices $v$ and $w$ in $\mathcal{K}$, if $b^t_v = b^t_w$, then $a^t_v = a^t_w$. We proceed by induction on $t$.
 
\textit{1) Base Case:} At $t = 0$, all simplices are assigned the same initial color, so the base case trivially holds.

\textit{2) Inductive Step:} Suppose that for some iteration $t$, $b^{t+1}_v = b^{t+1}_w$ for two vertices $v$ and $w$ in two arbitrary directed simplicial complexes $\mathcal{K}_1$ and $\mathcal{K}_2$, meaning that the D-SWL colourings of $v$ and $w$ at step $t+1$ are identical. Since vertices have no boundary simplices and are only upper adjacent, this implies: \[
b^t_v = b^t_w, \quad (b^{01}_{\uparrow})^t(v) = (b^{01}_{\uparrow})^t(w), \quad (b^{10}_{\uparrow})^t(v) = (b^{10}_{\uparrow})^t(w).
\]
Expanding the definition of these multisets, we get:\[
\{\{b^t_z \mid (b^t_z, \cdot) \in (b^{01}_{\uparrow})^t(v)\}\} = \{\{b^t_u \mid (b^t_u, \cdot) \in (b^{01}_{\uparrow})^t(w)\}\},
\]
\[
\{\{b^t_z \mid (b^t_z, \cdot) \in (b^{10}_{\uparrow})^t(v)\}\} = \{\{b^t_u \mid (b^t_u, \cdot) \in (b^{10}_{\uparrow})^t(w)\}\}.
\]
These are equivalent to the upper in- and out-neighborhood sets of the vertices $v$ and $w$, defined in \cite{Rossi23} as $N_{\leftarrow}(v)$ and $N_{\rightarrow}(v)$, respectively. Thus, we can rewrite the expressions as:
\[
\{\{b^t_z \mid z \in N_{\leftarrow}(v)\}\} = \{\{b^t_u \mid u \in N_{\leftarrow}(w)\}\},
\]
\[
\{\{b^t_z \mid z \in N_{\rightarrow}(v)\}\} = \{\{b^t_u \mid u \in N_{\rightarrow}(w)\}\}.
\] By the induction hypothesis, we know that $a^t_v = a^t_w$, $(a^{01}_{\uparrow})^t(v) = (a^{01}_{\uparrow})^t(w)$. Additionally, by Lemma 2 \cite{bevilacqua2021equivariant}, we have $(a^{10}_{\uparrow})^t(v) = (a^{10}_{\uparrow})^t(w)$. These are the arguments that the D-WL hash function uses to compute the colours of $v$ and $w$ at the next iteration, which implies that $a^{t+1}_v = a^{t+1}_w$. Hence, $b^t$ refines $a^t$, proving that D-SWL is at least as expressive as D-WL.
\subsection{Strictly More Expressive}
To demonstrate that D-SWL is strictly more powerful than D-WL, we present a counterexample: a pair of digraphs that cannot be distinguished by D-WL but whose directed flag complexes can be distinguished by D-SWL. In Fig. \ref{fig:dswl}, we show such a pair of digraphs. While the D-WL test produces identical colourings for both digraphs, their directed flag complexes reveal a crucial difference: one complex contains directed triangles, while the other contains none. Since D-SWL detects these higher-order simplicial structures, it successfully distinguishes the two digraphs, completing the proof.
\end{proof}

\textbf{Dir-SNNs vs Dir-GNNs}. We can finally prove Thm. 1 from Sec. \ref{sec:spnns}.

\begin{theorem} 
There exist Dir-SNNs that are more powerful than Dir-GNNs \cite{Rossi23} at distinguishing non-isomorphic digraphs when using a directed flag complex lifting.
\end{theorem}

\begin{proof}
First, we demonstrate that Dir-SNNs with sufficient layers and injective neighborhood aggregators are as powerful as the D-SWL test in distinguishing non-isomorphic directed simplicial complexes. The proof parallels the structure of Lemma 9 in \cite{bodnar2021weisfeiler}. \setcounter{subsection}{0}\subsection{Dir-SNNs As Powerful As D-SWL}
Let $c^t$ and $h^t$ represent the colouring at iteration $t$ of D-SWL and the $t$-th layer of a Dir-SNN, respectively. We consider a Dir-SNN with $L$ layers, and for $t > L$, we assume that $h^t = h^L$. First, we show by induction that the colouring $c^t$ of D-SWL refines the colouring $h^t$ of Dir-SNN. 

\subsubsection{Base Case}
At $t = 0$, all simplices are assigned the same initial color in both D-SWL and Dir-SNN, so the base case trivially holds.

\subsubsection{Inductive Step}
For the inductive step, suppose that for some iteration $t$, the colouring $c^{t+1}_\sigma = c^{t+1}_\tau$ for two simplices $\sigma$ and $\tau$ in a directed simplicial complex. Because the D-SWL colouring is injective, the arguments to the $\operatorname{HASH}$ function must also be equal. This implies the following equalities:

\[
c^t_\sigma = c^t_\tau, \quad c_{\mathcal{B}}^t(\sigma) = c_{\mathcal{B}}^t(\tau), \quad c_{\mathcal{C}}^t(\sigma) = c_{\mathcal{C}}^t(\tau),
\]
\[
((c_{\downarrow}^{ij})^t(\sigma))_{i,j} = ((c_{\downarrow}^{ij})^t(\tau))_{i,j}, \text{ and } ((c_{\uparrow}^{ij})^t(\sigma))_{i,j} = ((c_{\uparrow}^{ij})^t(\tau))_{i,j}.
\]
By the induction hypothesis, these arguments will also be equal under the Dir-SNN colouring $h^t$. Since the same arguments are supplied as input to the message-passing, aggregate, and update functions of Dir-SNN, their outputs will be identical for $\sigma$ and $\tau$. Therefore, $h_{t+1}(\sigma) = h_{t+1}(\tau)$. Following the discussion in Theorem 9 of \cite{bodnar2021weisfeiler}, and assuming that the boundary aggregation function is injective and \textit{non-permutation invariant}, we can derive the reverse implication: $h^t$ refines the colouring of $c^t$, proving that Dir-SNNs with sufficient layers and injective neighborhood aggregators are at least as powerful as D-SWL. 

\subsection{Strictly More Expressive}
Finally, applying Lemma \ref{thm:dswl-dwl}, we conclude that Dir-SNNs, when utilizing a directed flag complex lifting, are strictly more powerful than Dir-GNNs, as D-SWL is more expressive than D-WL in distinguishing non-isomorphic digraphs.
\end{proof}

\end{document}